%% file: main.tex
\documentclass{article}
\usepackage[]{hyperref}
\usepackage[preprint]{neurips_2024}
\usepackage{amsmath,amsfonts}
\usepackage[]{booktabs}
\usepackage{array}
\usepackage{textcomp}
\usepackage{url}
\usepackage{verbatim}
\usepackage{graphicx}
\usepackage[binary-units]{siunitx}
\usepackage{algorithm}
\usepackage{algpseudocode}
\usepackage{appendix}
\usepackage{subfig}

\title{Deep Reinforcement Learning for Time-Critical Wilderness Search And Rescue Using Drones}

\author{
  Jan-Hendrik Ewers\\Aerospace Sciences Research Division\\School of Engineering\\University of Glasgow\\\texttt{j.ewers.1@research.gla.ac.uk}
  \And
  David Anderson\\Aerospace Sciences Research Division\\School of Engineering\\University of Glasgow\\\texttt{dave.anderson@glasgow.ac.uk}
  \And
  Douglas Thomson\\Aerospace Sciences Research Division\\School of Engineering\\University of Glasgow\\\texttt{douglas.thomson@glasgow.ac.uk}
  }

\renewcommand{\vec}[1]{\mathbf{#1}}

\newcommand{\tech}[1]{\textit{#1}}

\renewcommand{\min}{\text{min}}
\renewcommand{\max}{\text{max}}

\DeclareSIUnit\km{\kilo\metre}
\DeclareSIUnit\kmsq{\km\squared}

\usepackage{xcolor}

\newcommand{\lhcgwconv}{LHC\_GW\_CONV\ }

\graphicspath{
  {sections/method/figures/}
  {sections/results/figures/}
  {sections/appendix/figures/}
}

\newif\ifdraft\draftfalse

\begin{document}

\maketitle

\input{sections/abstract/abtract.tex}

  \ifdraft  
    \include{sections/intro/introduction}
  \else
    \input{sections/intro/introduction}
  \fi

  \ifdraft  
    \include{sections/related_work/related_work}
  \else
    \input{sections/related_work/related_work}
  \fi

  \ifdraft  
    \include{sections/method/method}

  \else
    \input{sections/method/method}
  \fi

  \ifdraft  
    \include{sections/results/results}
  \else
    \input{sections/results/results}
  \fi

  \ifdraft  
    \include{sections/limitations}
  \else
    \input{sections/limitations}
  \fi

  \ifdraft  
    \include{sections/conclusion/conclusion}
  \else
    \input{sections/conclusion/conclusion}
  \fi

\bibliography{references}

  \ifdraft  
    \include{sections/appendix/appendix}
  \else
    \input{sections/appendix/appendix}
  \fi

% \inp{checklist}

\end{document}

%% file: sections/abstract/abtract.tex
\begin{abstract}
	Traditional search and rescue methods in wilderness areas can be time-consuming and have limited coverage. 
	Drones offer a faster and more flexible solution,
	but optimizing their search paths is crucial. 
	This paper explores the use of deep reinforcement learning to create efficient search missions for drones in wilderness environments.
	Our approach leverages a priori data about the search area and the missing person in the form of a probability distribution map. 
	This allows the deep reinforcement learning agent to learn optimal flight paths that maximize the probability of finding the missing person quickly.  
	Experimental results show that our method achieves a significant improvement in search times compared to traditional coverage planning and search planning algorithms. 
	In one comparison,
	deep reinforcement learning is found to outperform other algorithms by over $160\%$,
	a difference that can mean life or death in real-world search operations
	Additionally,
	unlike previous work,
	our approach incorporates a continuous action space enabled by cubature,
	allowing for more nuanced flight patterns.
\end{abstract}

%% file: sections/intro/introduction.tex
\section{Introduction}
\label{sect:intro}

Wilderness Search and Rescue (WiSAR) operations in Scotland's vast and often treacherous wilderness pose significant challenges for emergency responders.
To combat this,
Police Scotland Air Support Unit (PSASU) and Scottish Mountain Rescue (SMR) regularly use helicopters to assist in search operations
\cite{carrell_flying_2022}.
However,
the deployment of helicopters can be slow,
especially in the Scottish Isles where the PSASU's Eurocopter EC135 based in Glasgow can take multiple hours to arrive.
Additionally,
operating helicopters is extremely costly.

Drones, 
also known as unmanned aerial vehicles,
offer a cost-effective and agile solution for aerial search.
Both PSASU and SMR are placing small fleets of drones around Scotland for rapid deployment in a WiSAR scenario.
These fleets will never replace the requirement for a helicopter due to the inherent lifting disparities between the two platforms but will ensure that the search can begin as soon as possible.
Successfully utilizing drones in WiSAR requires careful planning
\cite{skeleton_remotely_2020}.

The current approach to flying drones for PSASU is the pilot-observer model where two personnel are required at a minimum per drone.
In this setup,
the observer is in charge of maintaining visual line of sight at all times whilst the pilot can fly the drone and inspect the live camera feed.
\citet{koester_sweep_2004}
identifies that a foot-based searcher has a higher detection rate when not in motion,
similar to the behaviour exhibited by pilots who will fly to a location then stop and pan the camera
\cite{ewers_gis_2023}.
The cognitive load between in-motion and static search is evidently a barrier,
as having the drone in motion at all times will lead to more area coverage which could lead to saved lives.

Another class of algorithm that has surpassed human abilities in highly complex tasks is Deep Reinforcement Learning (DRL),
a subset of machine learning.
DRL has been successfully applied to playing video games
\cite{mnih_humanlevel_2015,openai_dota_2019}
and has achieved championship-level drone racing
\cite{kaufmann_championlevel_2023}.
The ability of DRL to generalize the problem and make decisions based on extensive training allows it to provide unique solutions that other algorithms may not be able to achieve.

At the core of the problem,
WiSAR mission planning involves the effective use of a priori data,
such as the place last seen,
age,
and fitness levels of the missing person.
This information can be used to generate a Probability Distribution Map (PDM)
\cite{ewers_gis_2023,hashimoto_agentbased_2022,seric_lost_2021},
which describes the probability of detecting the missing person at a given location and informs the search stage of the mission.

Our research hypothesis is that by using DRL combined with a continuous PDM,
a more effective search path can be created.
The intent of using DRL is to allow better PDM exploration whilst the continuous PDM prevents undesired noisy rewards
\cite{guo_robust_2023,fox_taming_2017}
during training.
We benchmark our algorithm against work from 
\citet{lin_uav_2009},
and a standard lawnmower pattern
\cite{subramanian_probabilistic_2020}.

The contribution of this research to the field is the unique usage of a PDM as part of the observation space for the DRL agent.
Similar DRL algorithms by 
\citet{talha_autonomous_2022}
and 
\citet{peake_wilderness_2020}
explore the environment whilst searching and do not have the complete PDM at the beginning.
Another DRL search algorithm from 
\citet{ebrahimi_autonomous_2021}
localizes missing people using radio signal strength indexes which,
again,
does not have the a priori data present that our method does.
Furthermore,
this is the first search algorithm that uses a continuous PDM during the evaluation stage,
enabling a continuous action space whilst maintaining accurate probability integration
\cite{ewers_enhancing_2024}.

Further related work is discussed in \autoref{sect:related_work},
and the methodology is presented in \autoref{sect:method}.
Results are shown in \autoref{sect:results} with limitations being discussed in \autoref{sect:limitations}, and a conclusion is drawn in \autoref{sect:conclusion}.

%% file: sections/related_work/related_work.tex
\section{Related Work}
\label{sect:related_work}

% SAR Mission Planning and the need for DRL

Coverage planning algorithms have been around for decades\cite{galceran_survey_2013} in various forms with the most well-known,
and intuitive,
being the parallel swaths (also known as lawnmower or zig-zag) pattern.
This guarantees complete coverage of an entire area given enough time.
However,
for WiSAR applications,
reducing the time to find is substantially more important than searching the entire area.
In order to break away from the complete coverage problem definition,
a different objective needs to be defined.
This comes in the form of maximizing the accumulated probability over a PDM.
PDMs can be created in a number of ways such as through geometrical approaches\cite{heth_characteristics_1998},
agent-based methods\cite{hashimoto_agentbased_2022},
or using machine learning\cite{seric_lost_2021}.
All of these attempt to model the probability of where a lost person is located using information such as where they were last seen,
age,
sex,
and more.
This is a powerful tool in narrowing down a larger search area into something more manageable.

\citet{lin_uav_2009} approach the search planning problem by using a gradient descent algorithm in the form of Local Hill Climbing (LHC) that can advance into any of the surrounding eight cells.
However,
LHC alone is not sufficient because \citet{waharte_supporting_2010} found that this class of algorithm does not perform well due to their propensity in getting stuck around local maxima.
For this reason \citet{lin_uav_2009} introduces the notion of global warming to break out of local maxima.
This raises the zero probability floor sequentially a number of times,
storing the paths and then reassessing them given the original PDM.
Through this,
and a convolution-based tie-breaking scheme,
\lhcgwconv (local hill climb,
global warming,
convolution) is shown to have very favourable results.
Yet,
at every time step only the adjacent areas are considered.
This means that areas of higher total probability may not ever be considered until the current one is entirely covered.

Another approach to overcome this problem is to use a dynamic programming-based approach.
\citet{snyder_path_2021} uses Djikstra's algorithm by identifying the points of highest probability on the PDM and applying the point-to-point path planning algorithm from the start,
to these points,
and then finally back to the start when the endurance limit is equal to the distance back.
The reward function along the route is the probability seen by advancing into that position.
Like \lhcgwconv,
this is an iterative algorithm allowing movement into one of the surrounding cells before repeating the process.
Whilst this approach maximises the probability on the route towards the global maxima,
this does not consider other areas which could yield more total probability.

In order to consider the area as a whole,
sampling-based optimisation approaches have been applied to the problem.
\citet{morin_ant_2023} uses ant colony optimisation with a discrete PDM and \citet{ewers_optimal_2023} uses both genetic algorithm and particle swarm optimisation with a pseudo-continuous PDM.
Both are capable of finding solutions that do not necessarily need to include the global maxima,
however due to the nature of sampling-based optimisation problems they are prone to long computation times to converge on a solution.

% Why SAC over PPO
% Feature extractor engineering (NatureCNN)
A core problem with the previously mentioned algorithms is the inability to consider the PDM as a whole when making decisions.
Being able to prioritise long-term goals over short-term gains is a key feature of DRL.
This strand of machine learning couples classic reinforcement learning and neural networks to learn the optimal action based on an observation.

DRL is being used extensively for mission planning such as by 
\citet{yuksek_cooperative_2021} who used proximal policy optimisation to create a trajectory for two drones to avoid no-fly-zones whilst tracking towards the mission objective.
This approach has defined start and target locations,
however the uses of no-fly-zones with constant radius is analogous to an inverted PDM.
\citet{peake_wilderness_2020} uses Recurrent-DDQN for target search in conjunction with A2C for region exploration with a single drone to find missing people.
This method does not use any a priori information but rather explores the area in real time.
However,
in the end it does not outperform parallel swaths by a significant margin as would be expected by a more complex solution.
This shows that DRL is a suitable approach to the search-over-PDM problem that a WiSAR mission requires.

A core aspect of DRL is having a fully observable environment such that the policy can infer why the action resulted in the reward.
This is known as the markov decision process and is critical.
Thus,
being able to represent the PDM effectively is a primary goal.
Whilst images can be used as inputs for DRL,
as done by \citet{mnih_humanlevel_2015} to play Atari 2600 computer games,
the typically large dimension can be prohibitive.
Being able to represent the same information in a more concise manner reduces the observation space size resulting in lower training overhead.
However,
since PDM generation algorithms typically create discrete maps\cite{seric_lost_2021,heth_characteristics_1998,hashimoto_agentbased_2022},
finding a different representation is required.
To go from a discrete to continuous PDM,
\citet{lin_hierarchical_2014} uses a gaussian mixture model to represent the PDM as a sum of bivariate Gaussians.
\citet{yao_optimal_2019} uses a similar approach to approximate a univariate Gaussian along a river for search planning.
This can be easily used to numerically represent the numerous bivariate Gaussian parameters in an array which is a suitable format for a DRL observation.

An additional benefit of having a continuous representation is the performance increase,
in both noise and speed,
of calculating the accumulated probability as evaluated by \citet{ewers_enhancing_2024}.
It was found that the sampling-based approach typically used in conjunction with discrete PDMs only resulted in better speeds at low dimensions compared to the cubature method\cite{cools_algorithm_1997}.
It was also found that the cubature method always resulted in lower noise in the test cases where the sampling-based method was faster.
This is beneficial to ML in general,
as higher quality training data is always desired.

There are many DRL algorithms to chose from with proximal policy optimisation\cite{schulman_proximal_2017} and Soft Actor-Critic (SAC)\cite{haarnoja_soft_2018} being some of the most prevalent in the literature\cite{yuksek_cooperative_2021,mock_comparison_2023,xu_learning_2019}.
\citet{mock_comparison_2023} found that proximal policy optimisation performed well for low dimension observation spaces,
whilst SAC performed much better for larger ones.
The need for a large observation space comes from the fact that the policy would need have a sense of memory regarding where it had been  to encounter \tech{unseen} probability to satisfy the markov decision process that underpins DRL.
\citet{mock_comparison_2023} found that a recurrent architecture was comparable to including previous states in the observation (also known as frame stacking).
This shows that frame-stacking with SAC is a suitable DRL architecture for the current problem.

%% file: sections/method/method.tex
\section{Method}
\label{sect:method}
\newcommand{\ngauss}{{N_\textit{gaussian}}}
\newcommand{\rbuffer}{{R_\textit{buffer}}}
% Talk about the setup (environment -> model -> action -> ...) here

All parameters used in this study can be found in \autoref{sect:parameters}.

\subsection{Modelling}

\input{sections/method/agent_dynamics.tex}

\input{sections/method/pdm_modelling.tex}

\input{sections/method/reward_modelling.tex}

\input{sections/method/drl.tex}

%% file: sections/method/agent_dynamics.tex
\subsubsection{Environment}
\label{sect:method_agent_modelling}

The drone within the environment is modelled as a simple heading control model with a constant step size $\lambda$.
It is assumed that any drone executing this mission can perfectly track the waypoints through its controller or operator. 
Thus, the position vector $\vec x \in \mathbb{R}^2$ is updated via
\begin{gather}
	\vec x_{t+1} = \vec x_t +
	\lambda\begin{bmatrix}
	\cos u_t \\
	\sin u_t
	\end{bmatrix} \\
	u_t = \pi (a_t+1)
\end{gather}
where $a_t \in [-1,1]$ is the policy action at time-step $t$.

%% file: sections/method/pdm_modelling.tex
\subsubsection{PDM}
\label{sect:method_pdm_modelling}

\begin{figure}[htbp]
	\centering
	\includegraphics[width=0.5\textwidth]{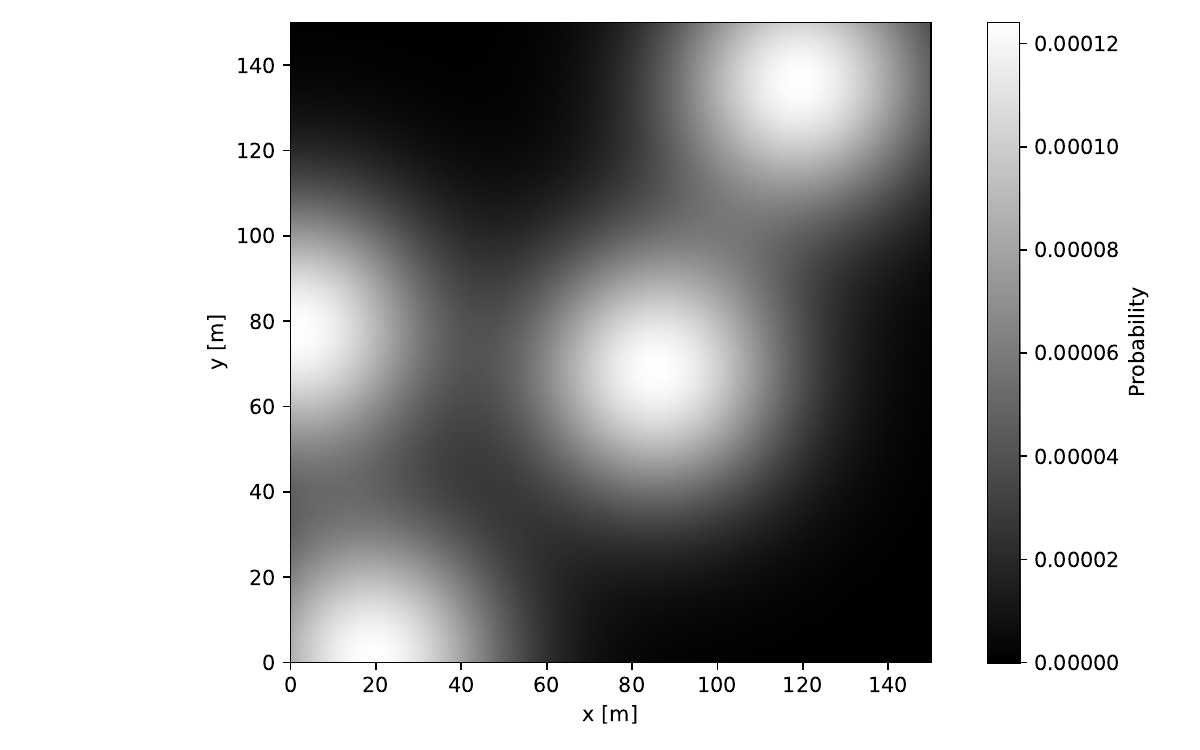}
	\caption{An example multi modal bivariate Gaussian PDM}
	\label{fig:example_pdm}
\end{figure}

The PDM is modelled as a sum of $\ngauss$ bivariate Gaussians\cite{yao_optimal_2019} such that a point on the ground at coordinate
$\vec x \in \mathbb R^2$
has a probability of containing the missing person
\begin{gather}
	p(\vec x) =
	\frac{1}{\ngauss} 
	\sum^\ngauss_{i=0}
	\frac
	{
		\exp{
			\left[
				-\frac
				{1}
				{2}
				(\vec x - \vec \mu_i )^T\vec\sigma_i^{-1}(\vec x - \vec \mu_i) 
			\right]
		}
	}
	{\sqrt{4\pi^2\det{\vec\sigma_i}}}
	\label{eqn:sum_of_bivariate_gaussians} \\
	\forall i \in [0,G],
	\vec \mu_i \sim \mathcal{U}([x_\min , x_\max],
	[y_\min, y_\max]
	)
\end{gather}
where $\vec \mu_i$
and 
$\vec \sigma_i$ 
are the mean location and covariance matrix of the $i$th bivariate Gaussian respectively.
If the bounding area were infinite, that is 
$x_\min = y_\min = -\infty \si\meter$ 
and 
$x_\max = y_\max = \infty \si\meter$,
then $\sum p(\vec x) = 1$. 
However, 
as can be seen from \autoref{fig:example_pdm}, 
the area enclosed by the rectangular bounds contain less than this. 
\autoref{sect:reward_modelling} further discusses how this is handled such that the available probability is normalized.

%% file: sections/method/reward_modelling.tex
\subsubsection{Reward}
\label{sect:reward_modelling}

\begin{figure}[htbp]
	\centering
	\subfloat[
		An example of how the buffered path polygon $h$ automatically deals with re-seen areas. Note that the highlighted areas are just for demonstration and are not a part of the algorithm.
		\label{fig:buffered_linestring_with_overlap}
	]{
		\includegraphics[width=0.38\linewidth]{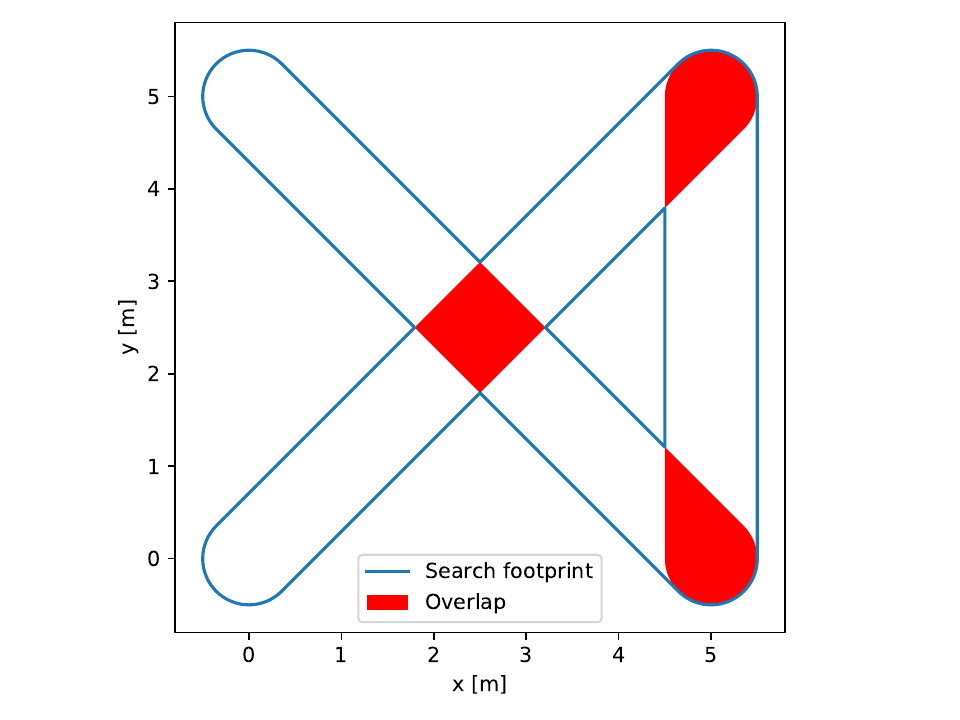}
	}
	\hfill
	\subfloat[
		The polygon from \autoref{fig:buffered_linestring_with_overlap} triangulated using the Delauney constrained triangulation.
		\label{fig:buffered_linestring_triangulated}
	]{
		\includegraphics[width=0.38\linewidth]{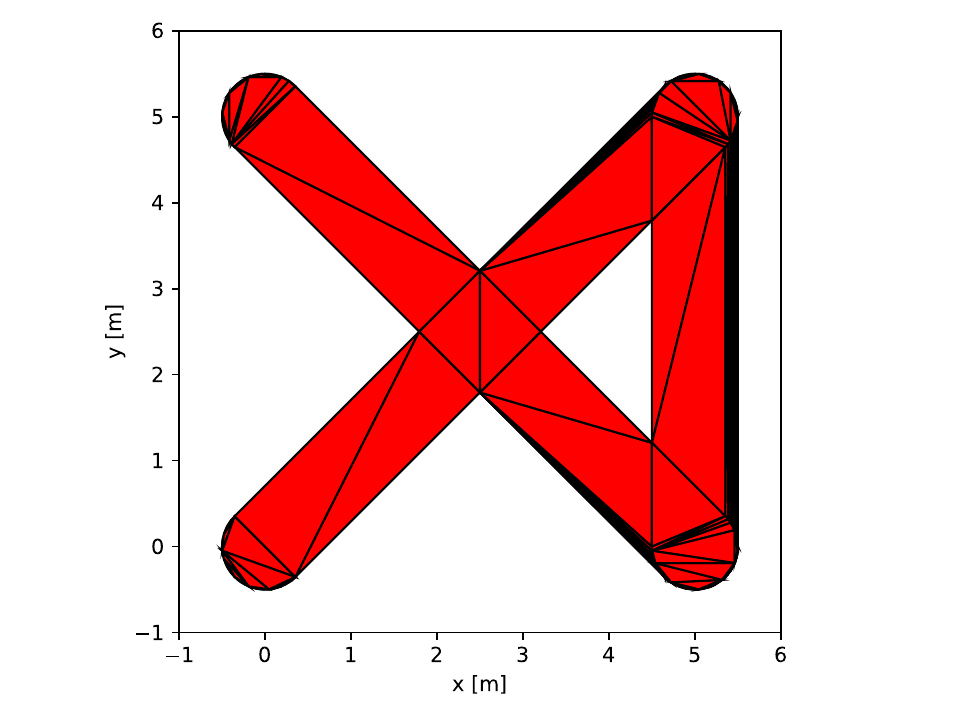}
	}
	\caption{Visualizations of concepts related to the buffered polygon representation of the seen area.}
	\label{fig:buffered_linestring}
\end{figure}

As the agent moves a constant distance $s\si\meter$ every step, it is assumed that the camera follows this path continuously at a fixed height whilst pointing straight down at all times. 
Therefore, to represent the \tech{seen area} for a given path at time-step $t$, the path is buffered by $\rbuffer\si\meter$ to give the polygon $h_t$. 
All probability from the PDM enclosed within $h_t$ is then \tech{seen} and denoted by $p_t$. 
This value, the seen probability, is calculated through
\begin{equation}
	I(H) = \int_H p(\vec x) dH
\end{equation}
with $H=h_t$. 
$p(\vec x)$ is from \autoref{eqn:sum_of_bivariate_gaussians}. 
Thusly, 
\begin{equation}
	p_t = I(h_t)
	\label{eqn:accumulated_probability_at_t}
\end{equation}
The integral is calculated using a cubature integration scheme\cite{ewers_enhancing_2024,cools_algorithm_1997} with constrained Delaunay triangulation\cite{chew_constrained_2023} to subdivide $H$ into triangles as seen in \autoref{fig:buffered_linestring_triangulated}. 

Other than allowing easy calculation of the accumulated probability, the buffering of the path prevents revisiting of an area contributing the same probability multiple times. 
This can be seen at the cross-over point $(2.5,2.5)\si\meter$ in \autoref{fig:buffered_linestring_with_overlap}.
											
\begin{figure}
	\centering
	\def\svgwidth{0.45\textwidth}
	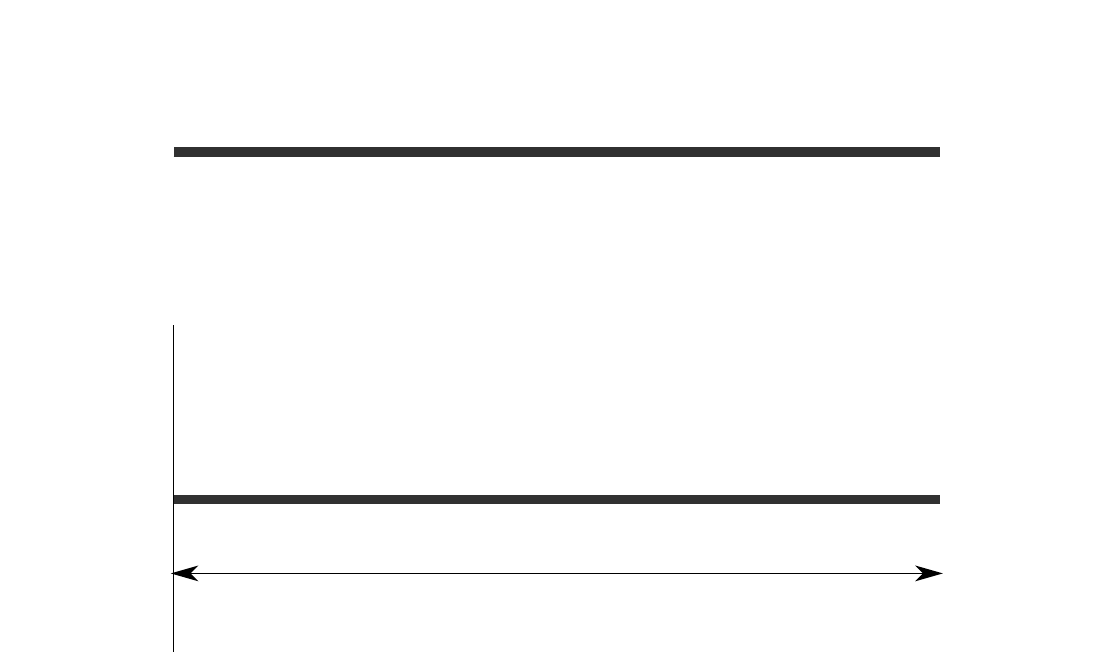
	\caption{The anatomy of the area calculation of an isolated step of length $\lambda$ with buffer $\rbuffer$ as used in \autoref{eqn:k}.}
	\label{fig:area_of_a_step}
\end{figure}

In order to correlate action to reward, only the additional probability that has been accumulated 
\begin{equation}
	\Delta p_t = p_t - p_{t-1}\label{eqn:delta_p_t}
\end{equation} 
is used.									
To normalize this value, the scaling constant $k$ is introduced. 
This scales $\Delta p_t$ by the ratio of the area of an isolated step $d\si\meter$ to the area of the total search area $a\si{\meter^2}$. 
This is defined as
\begin{align}
	k & = \frac{a_\textit{area}}{2 (\frac{1}{2} \pi \rbuffer^2 ) + 2 \rbuffer \lambda} \\
	  & = \frac{a_\textit{area}}{\rbuffer(\pi\rbuffer+2\lambda)}                       
	\label{eqn:k}            
\end{align}
with further spatial definitions from \autoref{fig:area_of_a_step}.
											
As highlighted in \autoref{sect:method_pdm_modelling}, the enclosed probability by the bounds is not equal to $1$.
To handle this, $\Delta p_t$ is scaled by the total available probability within the search area $p_A = I(A)$.
Combining $p_A$ with \autoref{eqn:delta_p_t} and \autoref{eqn:k}, gives the reward
\begin{equation}
	r = \frac{k}{p_A} \Delta p_t 
\end{equation}
The enclosed probability can then be used to calculate the probability efficiency at time-step $t$ with
\begin{equation}
	e_{p,t} =\frac{p_t}{p_A}
	\label{eqn:probability_efficiency}
\end{equation}
and $e_{p,t} \leq 1$.
											
Finally, reward shaping is used to discourage future out-of-bounds actions and to penalize visiting areas of low probability or revisiting previously seen sections. 
The latter is easily handled by the buffering of the path as seen in \autoref{fig:buffered_linestring_with_overlap}, 
where the areas highlighted in red will contribute nothing to the reward resulting in a penalty of $-w_{oob}$.
The augmented reward $r'$ is then defined as
% Reward
\begin{equation}
	r' = 
	\begin{cases}
		-w_{oob}, & \vec x_t \notin [x_\min,x_\max] \times [y_\min,y_\max] \\
		w_r r ,   & \Delta p_t > \epsilon                                  \\
		-w_{0}	,  & \textit{else}                                          
	\end{cases}
	\label{eqn:reward_with_cases}
\end{equation}

%% file: sections/method/figures/k.pdf_tex
%% Creator: Inkscape 1.3.2 (091e20ef0f, 2023-11-25), www.inkscape.org
%% Accompanies image file 'k.pdf' (pdf, eps, ps)
\begingroup%
  \makeatletter%
  \providecommand\color[2][]{%
    \errmessage{(Inkscape) Color is used for the text in Inkscape, but the package 'color.sty' is not loaded}%
    \renewcommand\color[2][]{}%
  }%
  \providecommand\transparent[1]{%
    \errmessage{(Inkscape) Transparency is used (non-zero) for the text in Inkscape, but the package 'transparent.sty' is not loaded}%
    \renewcommand\transparent[1]{}%
  }%
  \providecommand\rotatebox[2]{#2}%
  \newcommand*\fsize{\dimexpr\f@size pt\relax}%
  \newcommand*\lineheight[1]{\fontsize{\fsize}{#1\fsize}\selectfont}%
  \ifx\svgwidth\undefined%
    \setlength{\unitlength}{534.70735805bp}%
    \ifx\svgscale\undefined%
      \relax%
    \else%
      \setlength{\unitlength}{\unitlength * \real{\svgscale}}%
    \fi%
  \else%
    \setlength{\unitlength}{\svgwidth}%
  \fi%
  \global\let\svgwidth\undefined%
  \global\let\svgscale\undefined%
  \makeatother%
  \begin{picture}(1,0.58487605)%
    \lineheight{1}%
    \setlength\tabcolsep{0pt}%
    \put(0,0){\includegraphics[width=\unitlength,page=1]{k.pdf}}%
    \put(0.45898651,0.02269686){\color[rgb]{0.12941176,0.12941176,0.12941176}\makebox(0,0)[lt]{\lineheight{1.25}\smash{\begin{tabular}[t]{l}\textbf{$\lambda$}\end{tabular}}}}%
    \put(0,0){\includegraphics[width=\unitlength,page=2]{k.pdf}}%
    \put(0.00879669,0.53165457){\color[rgb]{0.12941176,0.12941176,0.12941176}\makebox(0,0)[lt]{\lineheight{1.25}\smash{\begin{tabular}[t]{l}\textbf{$\rbuffer$}\end{tabular}}}}%
  \end{picture}%
\endgroup%

%% file: sections/method/drl.tex
\subsection{Training Algorithm}

\begin{figure}[htbp]
	\centering
	\includegraphics[width=0.35\linewidth]{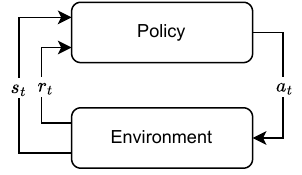}
	\caption{Top-level representation of a typical reinforcement learning data flow. The agent is also commonly referred to as the policy}
	\label{fig:rl_flow_diagram}
\end{figure}
	 
SAC\cite{haarnoja_soft_2018} is a DRL algorithm well-suited for continuous control problems like drone navigation\cite{kaufmann_championlevel_2023}. 
It tackles the challenge of exploration vs exploitation by simultaneously maximizing expected reward and entropy. 
Entropy encourages exploration of the environment, 
preventing the policy from getting stuck in local optima. 
SAC achieves this by learning a policy, 
multiple Q-functions to evaluate actions, 
and a temperature parameter\cite{haarnoja_soft_2019} that controls the trade-off between exploration and exploitation.
\autoref{fig:rl_flow_diagram} outlines the iterative process of reinforcement learning, where the policy is used to generate actions $a_t$ and the environment then provides the resultant observation $s_t$ and reward $r_t$.

\subsubsection{Policy Architecture}

The the core of the policy is compromised of a fully connected network with $N_\text{layers}$ layers, each with a width of $N_\text{width}$. 
This is given an input from the Feature Extractor (FE) which is further subdivided into five sub-FEs; 
path history $s_\text{path}$, PDM parameters $s_\text{PDM}$, out-of-bounds $s_\text{oob}$, and number of steps $s_\text{steps}$. 
The corresponding state observations for each of the sub-FEs is defined in \autoref{tbl:observation_states} and each observation is linearly normalized.
This results in a total of $2N_\text{waypoints}+6G+4$ observation inputs.
\newcommand{\spdm}{
	\left[		
		\vec\mu_0,
		\vec\sigma_0,
		\dots,
		\vec\mu_G,
		\vec\sigma_G 
	\right]^T
}
\newcommand{\spath}{
	\left( \left[ \vec x_0, \dots, \vec x_t \right]~\|~\mathbf{0}^{2\times N_\text{waypoints}-t} \right)^T
}
\begin{table}[htb]
	\centering
	\caption{Definition of the five state observations}
	\label{tbl:observation_states}
	\begin{tabular}{@{}llll@{}}
		\toprule
		Sub-state       & Symbol           & Definition                                            & Shape                   \\ \midrule
		Path            & $s_\text{path}$  & $\spath$                                              & $(2,N_\text{waypoint})$ \\
		PDM             & $s_\text{PDM}$   & $\spdm$                                               & $(2,3G)$                \\
		Position        & $s_\text{pos}$   & $\vec x_t$                                            & $(1,2)$                 \\
		Out-of-bounds   & $s_\text{oob}$   & $\vec x_t \in [x_\min,x_\max] \times [y_\min,y_\max]$ & $(1,)$                  \\
		Number of steps & $s_\text{steps}$ & $t$                                                   & $(1,)$                  \\ \bottomrule
	\end{tabular}
\end{table}

\begin{figure}[htbp]
	\centering
	\includegraphics[width=0.7\linewidth]{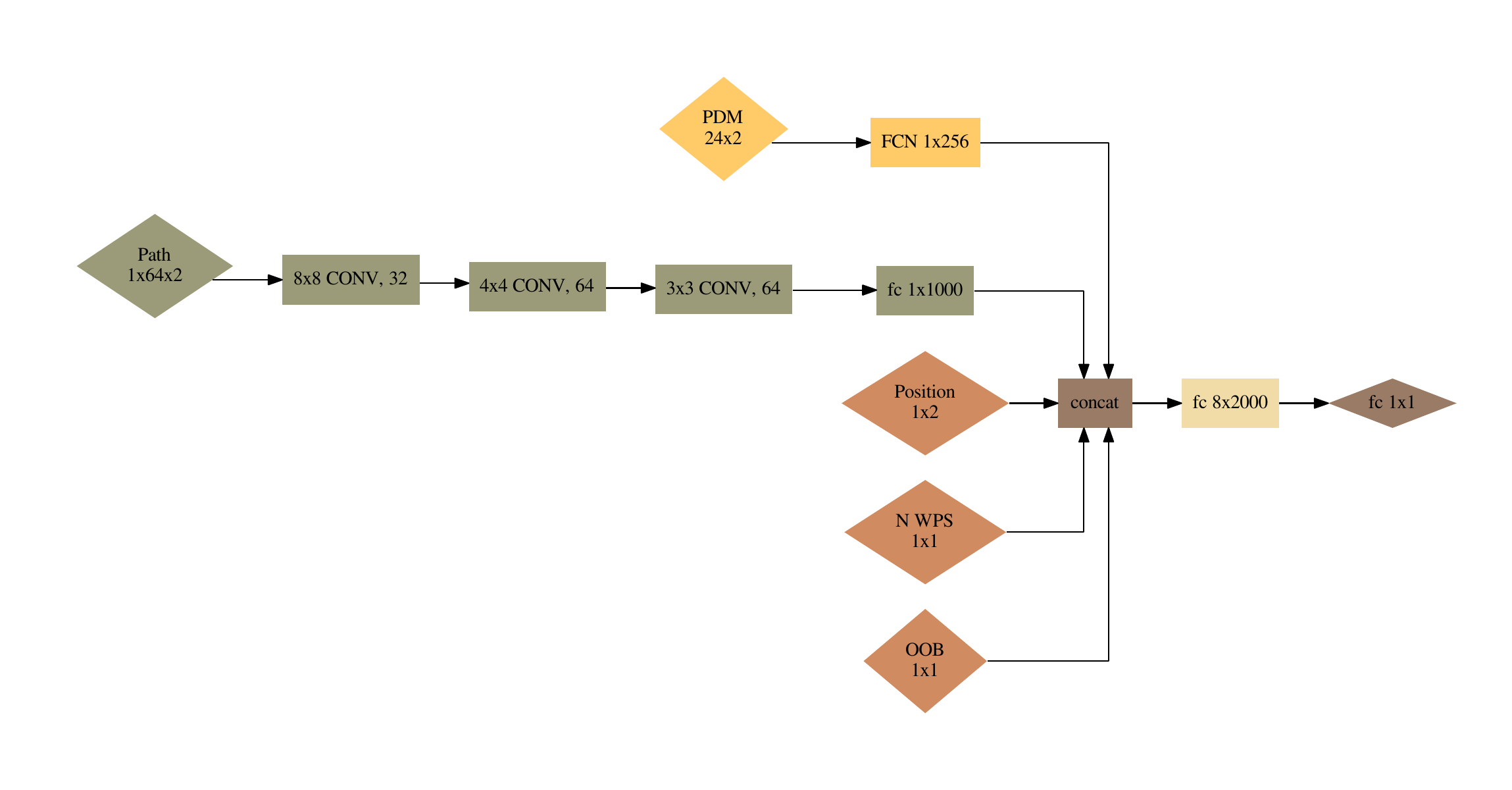}
	\caption{Policy network architecture}
	\label{fig:policy_net_arch}
\end{figure}

The policy architecture designed to handle this observation is defined in \autoref{fig:policy_net_arch}. The sub-FE for the $r_\text{path}$ state observation is of particular note as it uses a 2D CNN architecture based on work by \citet{mnih_humanlevel_2015}.
										
% Net arch design via the parameter sweep

%% file: sections/results/results.tex
\section{Results}
\label{sect:results}
\newcommand{\myalgorithm}{SAC-FS-CNN\ }
	
\subsection{Experimental setup}
\label{sect:results_experimental_setup}

In order to effectively benchmark the proposed algorithms, two additional baselines are implemented; lawnmower\cite{galceran_survey_2013} and \lhcgwconv\cite{lin_uav_2009}. 
These were chosen due to the former being ubiquitous for coverage planning, and the latter being a optimisation-based implementation that struggles to fully explore the PDM.
To ensure compatibility in the comparison to the proposed algorithm, the parallel lines for lawnmower are offset by the step size $\lambda$ and the grid dimensions for \lhcgwconv~are $(x_\max-x_\min, y_\max-y_\min)/\lambda$.
The maximum number of waypoints $N_\text{waypoint}$ are converted to a maximum distance $D_\max= \lambda N_\text{waypoint}$ and the generated paths are truncated at this point.

The results for the algorithm implemented in this research, titled \tech{\myalgorithm} from here on in, is the cumulation of three separate training runs with random starting seeds.
This aligns with the best practices outlined by \citet{agarwal_deep_2022} to ensure robust analysis for DRL results. 
Each model was trained for a minimum of $21$ days ($\num{5e8}$ global steps) with $32$ workers on a local \texttt{Ubuntu 22.04} machine with a
\texttt{AMD Ryzen 9 5950X} CPU with a \texttt{NVIDIA RTX A6000} GPU and $64\si{\giga\byte}$ of RAM.

One evaluation of an algorithm involves generating the random PDM, then creating the resultant search path. 
This is labelled one run. 
\autoref{tbl:n_runs} shows the number of runs undertaken per algorithm and this generated data is base of the following analysis.

\begin{table}[]
	\centering
	\caption{Number of runs per algorithm}
	\label{tbl:n_runs}
	\begin{tabular}{@{}lr@{}}
		\toprule
		Method     & $N$          \\
		\midrule
		\lhcgwconv & $\num{5e3}$  \\
		Lawnmower  & $\num{5e3}$  \\
		SAC-FS-CNN & $\num{10e3}$ \\
		\bottomrule
	\end{tabular}
\end{table}

\subsection{Probability Over Distance (POD)}
\label{sect:pod}

Maximising the probability efficiency (\autoref{eqn:probability_efficiency}) at all times is critical. 
This directly correlates to increasing the chances of finding a missing person in a shorter time. 
It is important for the POD of \myalgorithm to out-perform the benchmark algorithms at all times. 
If this is not the case, the search algorithm selection becomes dependent on the endurance and mission. However, if the POD is better at all times then one algorithm will be superior no matter the application.
To calculate the POD, the probability efficiency is evaluated at
\begin{equation}
	d = \frac{(N_\text{steps}-i)D}{N_\text{steps}} \forall i \in \{N_\text{steps},N_\text{steps}-1,\dots,1,0\}
	\label{eqn:path_walking}
\end{equation}
with $N_\text{steps} = 50$.
					
\begin{figure}[htbp]
	\centering
	\subfloat[
		Mean $e_p$ with $d$ showing \myalgorithm outperforming the benchmark algorithms at all distances.\label{fig:mean_p_efficiency_over_distance}
	]{
		\includegraphics[width=0.3\linewidth]{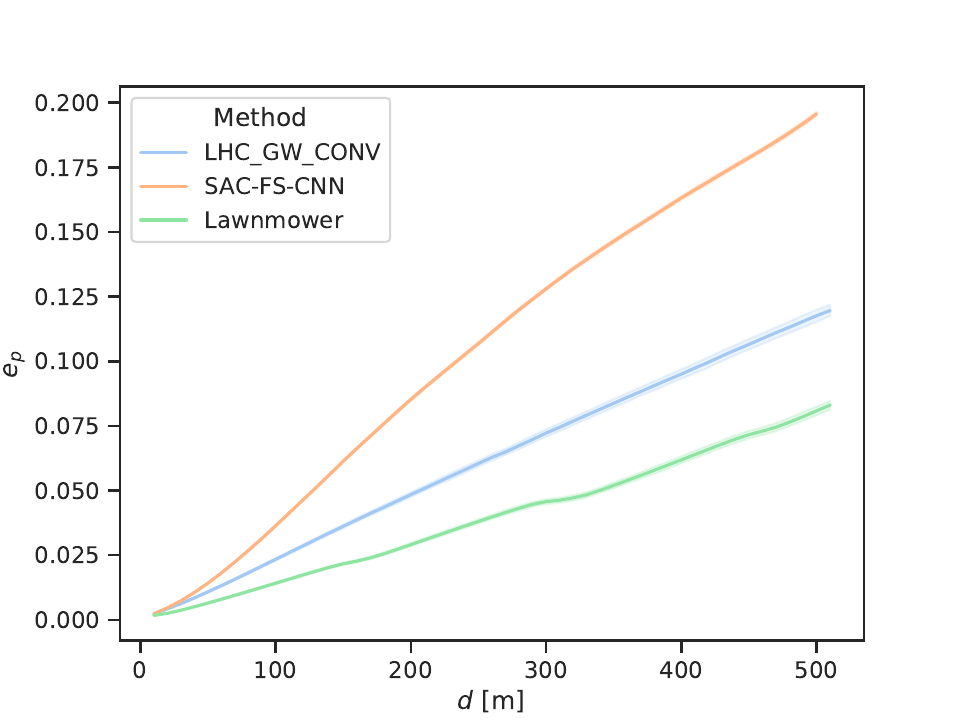}
	}
	\hfill
	\subfloat[Performance profile of $e_{p,D}$. \myalgorithm has a close to $100\%$ of runs with $e_{p,D}>0.1$ compared to the $<50\%$ for the other algorithms.
		\label{fig:perf_profile_p_efficiency}
	]{
		\includegraphics[width=0.3\linewidth]{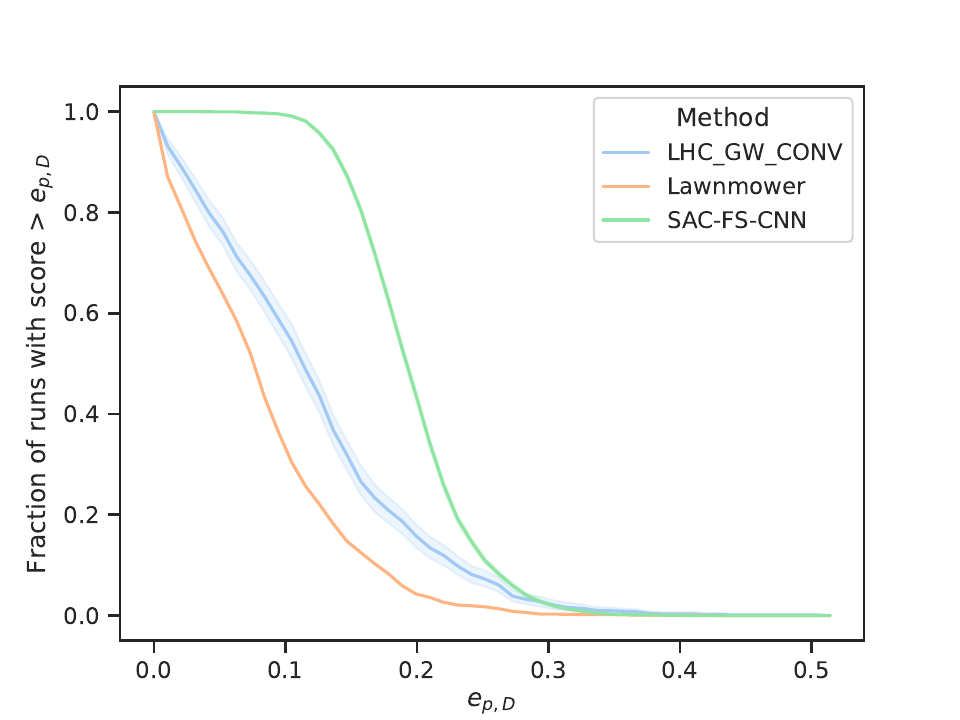}
	}
	\hfill
	\subfloat[Median $e_{p,D}$ for all runs. \myalgorithm has a higher median than the other algorithms. \lhcgwconv registers large outliers above $95\%$ highlighting the algorithms potential.
		\label{fig:final_p_efficiency_boxplot}
	]{
		\includegraphics[width=0.3\linewidth]{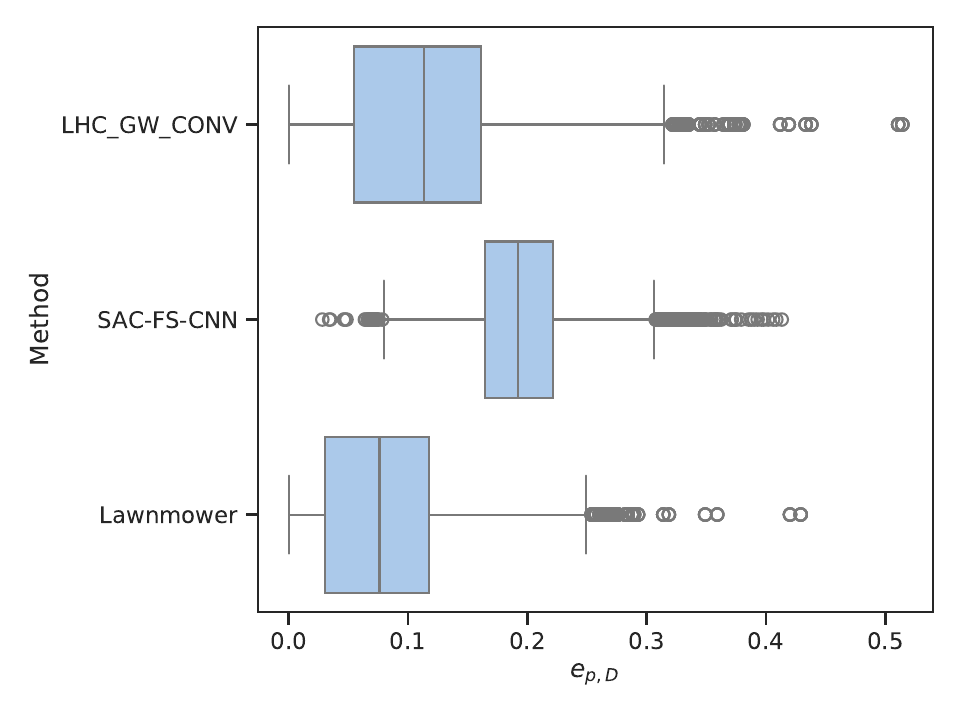}
	}
	\caption{Probability efficiency analysis over all runs. The shaded regions in \autoref{fig:mean_p_efficiency_over_distance} and \autoref{fig:perf_profile_p_efficiency} show the $95\%$ confidence interval.}
\end{figure}

\begin{table}[]
	\centering
	\caption{Mean POD performance metrics with the standard deviation as error.}
	\label{tbl:probabilities}
	\begin{tabular}{@{}lrrr@{}}
		\toprule
		Method     & $p_D$                  & $e_{p,D}$              & $N$           \\
		\midrule
		\lhcgwconv & $0.09\pm0.06$          & $0.12\pm0.08$          & $\num{9.3e3}$ \\
		Lawnmower  & $0.06\pm0.04$          & $0.08\pm0.06$          & $\num{10e3}$  \\
		SAC-FS-CNN & $\mathbf{0.15\pm0.04}$ & $\mathbf{0.19\pm0.04}$ & $\num{9.8e3}$ \\
		\bottomrule
	\end{tabular}
\end{table}

From \autoref{fig:mean_p_efficiency_over_distance} it is clear that \myalgorithm sufficiently outperforms the benchmark algorithms at all distances. This is further highlighted by the $e_{p,D}$ for \myalgorithm at $238\%$ of that of lawnmower, and $158\%$ for \lhcgwconv from \autoref{tbl:probabilities}. This is corroborated by the median $e_{p,D}$ values in \autoref{fig:final_p_efficiency_boxplot}. Notably, however, \lhcgwconv has a substantial amount of high $e_{p,D}$ outliers. 

Likewise, the performance profile from \autoref{fig:perf_profile_p_efficiency} follows the trend. It can be seen that \myalgorithm has close to $100\%$ of runs with $e_{p,D}>0.1$ and $50\%$ at approximately $e_{p,D} > 0.2$. This aligns with results from \autoref{fig:final_p_efficiency_boxplot} and \autoref{tbl:probabilities}.

\subsection{Distance To Find (DTF) and Percentage Found (PF)}

Whilst POD shows the theoretical effectiveness of an algorithm, the intended use-case is finding a missing person whilst searching within a bounded area. The mission statement is reducing the time it takes to find the potentially vulnerable person to save lives.

To quantify this requirement, we introduce DTF and PF. The former gives a clear answer on the capabilities on the various algorithms, whilst the latter should align with the POD results from \autoref{tbl:probabilities} for validation.

Firstly, Gumbel-Softmax\cite{li_localization_2021} is used to sample $N_\text{samples}$ positions from the PDM to give the set $\vec \chi \in \mathbb{R}^{2 \times N_\text{samples}}$ containing all samples. The path is then traversed in incremental steps using \autoref{eqn:path_walking} with $N_\text{step} = \num{1e4}$. At each step, a euclidean distance check is done from the current position $\vec x$ to each entry in $\vec\chi$ with any points within $\rbuffer$ being marked as seen. The updated set of positions to search for in the next step is then
\begin{equation}
	\vec\chi' = \{\vec\chi_i \in \vec\chi   :   \| \vec x - \vec\chi_i \| > \rbuffer\}
\end{equation}
	
\begin{figure}[htbp]
	\centering
	\includegraphics[width=0.45\linewidth]{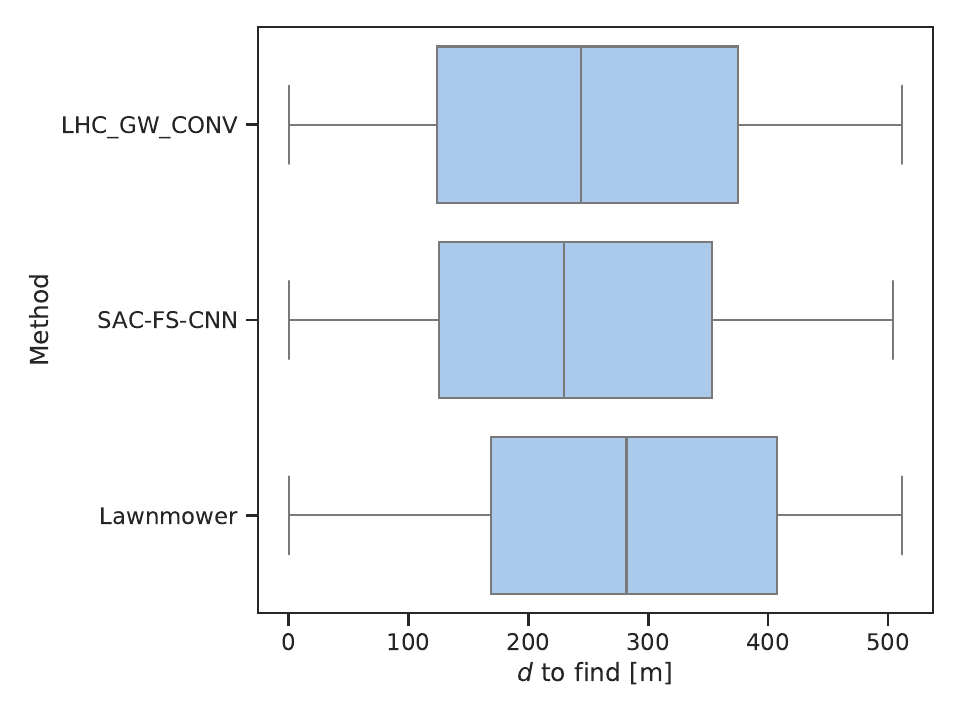} 
	\caption{Median $d$ to find. Similar $5\%$ and $95\%$ confidence markers across the algorithms shows the broad distribution of results from this experiment. \myalgorithm ultimately has the lowest median score.}
	\label{fig:distance_to_find}
\end{figure}
	
\begin{table}[]
	\centering
	\caption{DTF performance metrics}
	\label{tbl:distance_to_find}
	\begin{tabular}{@{}lrrr@{}}
		\toprule
		Method     & PF [$\%$]               & Mean DTF [$\si\meter$]     & $N$          \\ \midrule
		\lhcgwconv & $11.86\pm0.32$          & $249.37\pm145.70$          & $\num{5e6}$  \\
		Lawnmower  & $7.77\pm 0.27$          & $282.63\pm146.62$          & $\num{5e6}$  \\
		SAC-FS-CNN & $\mathbf{19.00\pm0.39}$ & $\mathbf{239.61\pm138.16}$ & $\num{10e6}$ \\ 
		\bottomrule
	\end{tabular}
\end{table}
	
From \autoref{fig:distance_to_find}, it is clear to see that \myalgorithm outperforms the benchmark algorithms with a lower median DTF as well as a lower inter-quartile range. \autoref{tbl:distance_to_find} shows that the mean DTF is $15.22\%$ lower than lawnmower, and $4.07\%$ lower than \lhcgwconv. This is in line with expectations from the results in \autoref{sect:pod}. Likewise, the PF values closely match to the $e_{p,D}$ values from \autoref{tbl:probabilities} showing that this test correlates to the theory.

%% file: sections/limitations.tex
\section{Limitations}
\label{sect:limitations}

DRL search mission planning holds promise,
but limitations exist. 
The proposed environment assumes a flat, confined world, 
hindering performance in complex environments with varying altitudes. 
Similarly, 
a limited camera view restricts the agent's ability to look around, 
as does the fixed height constraint.
Additionally, 
fixed waypoint spacing and difficulty scaling to different mission sizes reduce flexibility and adaptability.
Likewise, scaling to higher $\ngauss$ values, or other PDM representations also presents the same issues.  
Further research is needed to address these limitations and unlock DRL's fully for robust search mission planning in the real world.

%% file: sections/conclusion/conclusion.tex
\section{Conclusion}
\label{sect:conclusion}

Our research investigated the potential application of DRL for mission planning in WiSAR operations, leveraging a priori information.
This was identified as a solution to the challenge of maximizing accumulated probability over a given area due to the powerful capabilities of machine learning to identify patterns and make generalizations in complex tasks.

The results indicate that DRL can outperform benchmark algorithms in the probability efficiency by up to $250\%$ for lawnmower, 
and $166\%$ for \lhcgwconv. 
A similar trend is identified when comparing mean distance-to-find with DRL outperforming the aforementioned algorithms by $15.22\%$ and $4.07\%$ respectively. 
The critical result, 
however, 
was that DRL found $160\%$ more simulated missing people than \lhcgwconv.
This translates to a substantial advantage in locating missing individuals, potentially saving countless lives during WiSAR operations.

However,
alongside these promising advancements, 
ethical considerations regarding DRL-powered drones in WiSAR require careful consideration.
Unauthorized surveillance could lead to privacy violations,
while noise pollution might disrupt wilderness areas.
Furthermore, over-reliance on this technology might lead to overlooking vital details or malfunctions during critical rescue missions. 
By recognizing these potential drawbacks, we can ensure a more balanced and responsible approach to this promising technology.

Our research focused on simulations, 
highlighting the need for  further studies to evaluate DRL in real-world scenarios with more complex environmental factors.
Likewise, a more representative drone model should be tested to ensure compatibility between generated and physically feasible paths.
This step is vital to ensure a level of confidence that the generated paths do in fact have the capabilities to increase search performance. 
Without this, the generated paths could potentially hinder rather than assist fast and efficient WiSAR in the real world.

The integration of DRL into WiSAR mission planning holds great potential for the future of search,
offering a powerful tool with potential to significantly increase the success rate of WiSAR efforts.

%% file: sections/appendix/appendix.tex
\input{sections/appendix/parameters.tex}

\input{sections/appendix/policy_netork_design.tex}

%% file: sections/appendix/parameters.tex
\appendix

\section{Parameters}
\label{sect:parameters}

\begin{table}[htb]
	\centering
	\caption{Simulation parameters used for this study}
	\label{tbl:sim_parameters}
	\begin{tabular}{@{}lrl@{}}
		\toprule
		Parameter           & Value           & Units       \\ 
		\midrule
		$\ngauss$           & $4$             &             \\
		$\vec \sigma_i$     & $diag(500,500)$ &             \\
		$x_\min$, $y_\min$  & $0$             & $\si\meter$ \\
		$x_\max$, $y_\max$  & $150$           & $\si\meter$ \\
		$\lambda$           & $8  $           & $\si\meter$ \\
		$\rbuffer$          & $2.5$           & $\si\meter$ \\
		$N_\text{waypoint}$ & $64 $           &             \\
		$\epsilon$          & 0.1             &             \\
		$w_{oob}$ & $1.0$\\
		$w_r$ & $0.5$ \\
		$w_0$ & $0.5$\\
		\bottomrule
	\end{tabular}
\end{table}

\begin{table}[htb]
	\centering
	\caption{SAC hyperparameters used for this study from empirical testing. Other variables were kept at the default values from \citet{raffin_stablebaselines3_2021} \texttt{v2.1.0}}
	\label{tbl:sac_hyperparameters}
	\begin{tabular}{@{}lr@{}}
		\toprule
		Hyperparameter     & Value                       \\ 
		\midrule
		Learning rate      & $\num{1e-6}$                \\
		Optimizer          & Adam\cite{kingma_adam_2017} \\
		Batch size         & $1024$                      \\
		Learning starts    & $8192$                      \\
		Buffer size        & $\num{5e6}$                 \\
		Training frequency & $10$                        \\
		Gradient steps     & $50$                        \\
		$\tau$             & $\num{1e-4}$                \\ 
		\bottomrule
	\end{tabular}																								
\end{table}

%% file: sections/appendix/policy_netork_design.tex
\section{Policy Network Design}

Search planning is a abstract task compared to point-to-point or coverage planning.
Therefore,
a parameter sweep for the network architecture was performed to ensure sufficient capabilities.
A core aspect of this was the large path history observation with $2N_\text{waypoint}$ elements since
the policy must be able to learn how to avoid crossing over itself to avoid penalties.

In this sweep four variables were tuned; number of layers $N_\text{layers}$,
layer width $N_\text{width}$,
path output feature dimension,
and path feature extractor.
$N_\text{layers}$ and $N_\text{width}$ correlate to the core policy network,
and the other feature extractors are left unchanged and can be seen in 
\autoref{fig:policy_net_arch}.
The ranges of these sweeps can be seen from 
\autoref{fig:hyperparameter_sweep_ols}.
The SAC hyperparameters from \autoref{tbl:sac_hyperparameters} were used for all runs.

\begin{figure}[htbp]
	\centering
	\subfloat[Single-layer FCN path feature extractor\label{fig:path_fe_fcn}]{
		\includegraphics[height=8cm]{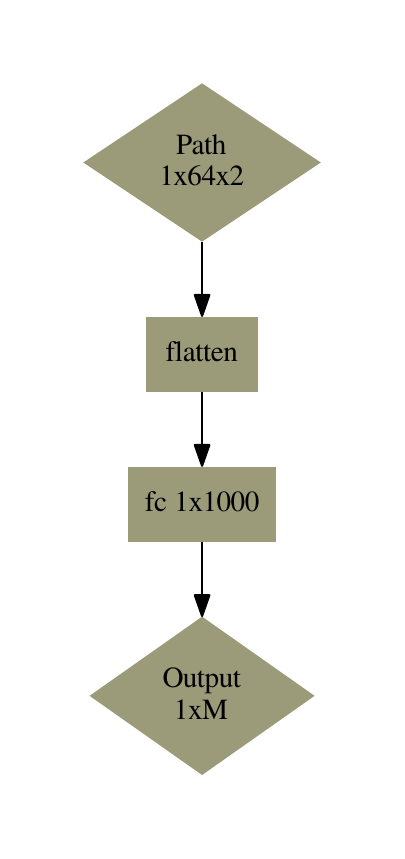}
	}
	\hfill
	\subfloat[1D convolution path feature extractor\label{fig:path_fe_conv1d}]{
		\includegraphics[height=8cm]{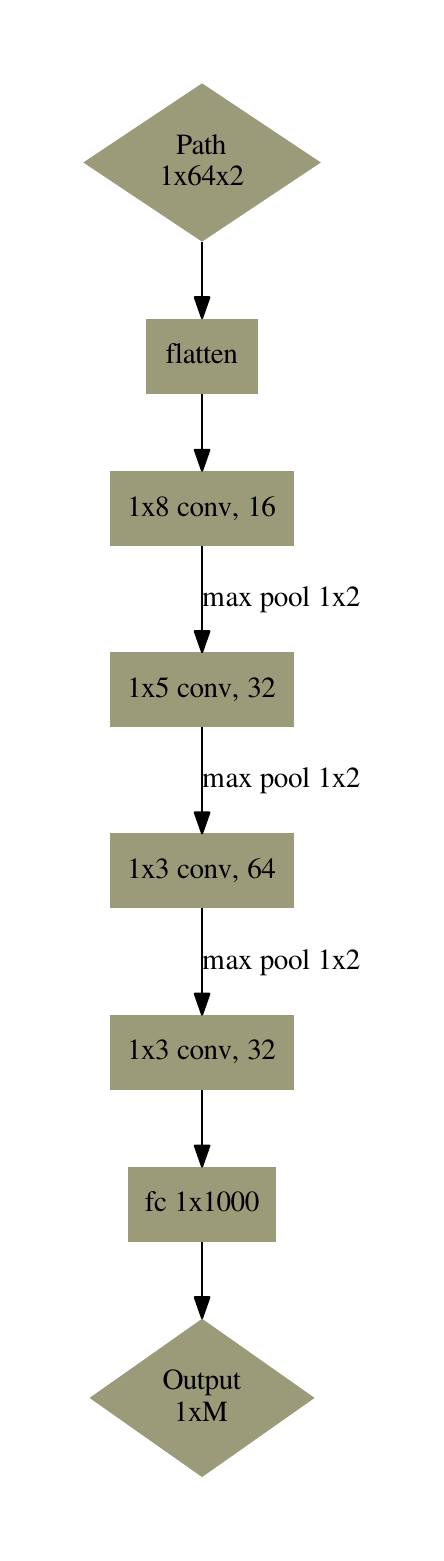}
	}
	\hfill
	\subfloat[2D convolution path feature extractor\label{fig:path_fe_conv2d}]{
		\includegraphics[height=8cm]{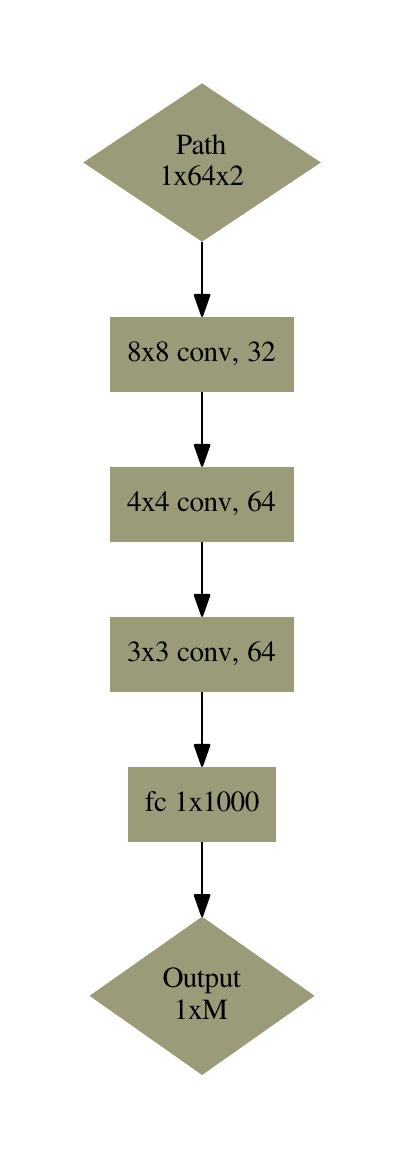}
	}
	\caption{Architectures of the three candidate path feature extractors. $M$ is the path output feature dimension.}
	\label{fig:path_fe}	
\end{figure}

There were three path feature extractors that were tested.
A standard FCN-based approach (
\autoref{fig:path_fe_fcn}),
a 2D CNN from 
\citet{mnih_humanlevel_2015} (
\autoref{fig:path_fe_conv2d}),
and a 1D CNN variant (
\autoref{fig:path_fe_conv1d}).

In total,
$39$ runs were completed over $12$ days in the same computational environment as outlined in \autoref{sect:results_experimental_setup}.
A random sampling strategy was used to ensure broad coverage of the hyperparameter space.
Results from this, with a fitted ordinary least squares (OLS) linear regression where applicable, can be seen in 
\autoref{fig:hyperparameter_sweep_ols}.

From the results in 
\autoref{fig:probability_efficiency_mean_vs_layer_width} and 
\autoref{fig:probability_efficiency_mean_vs_n_layers} it is evident that a deeper,
wider network leads to better results.
However,
\autoref{fig:probability_efficiency_mean_vs_n_layers} does taper off in performance after $N_\text{layers}=8$ with similar results.
The overall number of parameters per result were calculated and plotted in 
\autoref{fig:probability_efficiency_mean_vs_n_params}.
This confirms the result that a larger policy network is better.
A network size of $8\times2000$ was selected as seen in 
\autoref{fig:policy_net_arch}.

Results from the path output feature dimension in 
\autoref{fig:probability_efficiency_mean_vs_path_output_dim} would be expected to show similar results.
However,
the OLS has a downward trend implying lower values for being better.
Yet the fitted model did not exhibit statistically significant explanatory power,
with a p-value of $0.393$.
For comparison,
the p-values for $N_\text{layers}$ and $N_\text{width}$ were $0.00522$ and $0.00339$ respectively and are well below the standard test p-value of $0.05$.
Thus,
a path output feature dimension of $1000$ was selected.

Finally,
the result for the path feature extraction method in 
\autoref{fig:probability_efficiency_mean_vs_path_feature_extraction} highlight that the 2D CNN feature extractor yields the highest median $e_{p,D}$.
However,
the 1D CNN median result is very similar.
A p-value was calculated against each distribution to determine if the results were significantly different.

\begin{figure}[htbp]
	\centering
	\includegraphics[width=0.45\linewidth]{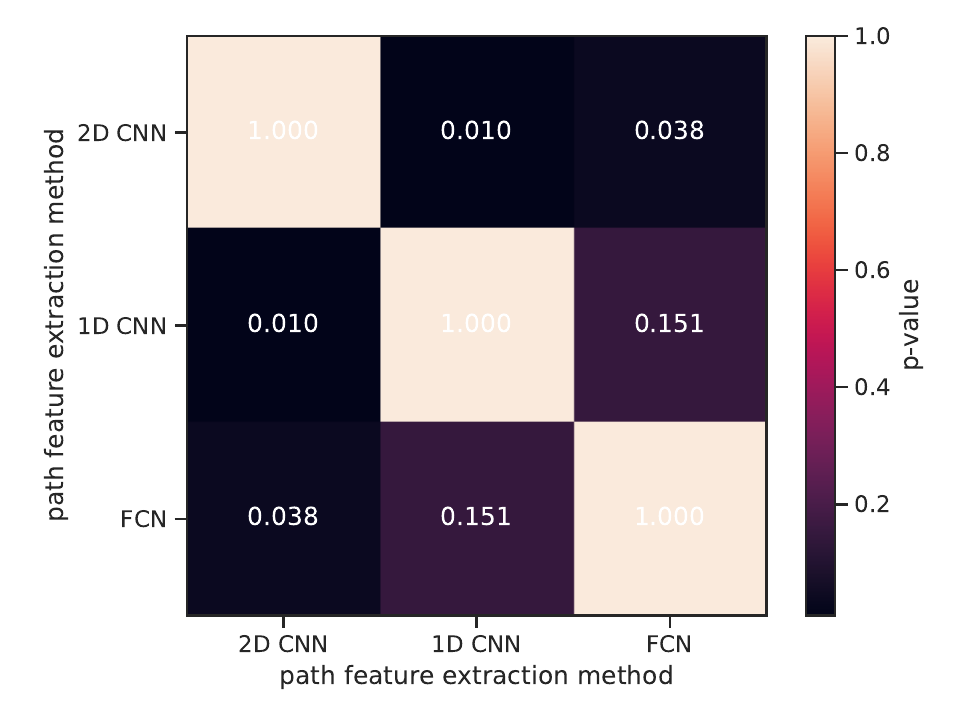}
	\caption{p-value matrix from comparing the result distributions categorized by feature extraction method.}
	\label{fig:hyperparameter_sweep_p_value}
\end{figure}

\autoref{fig:hyperparameter_sweep_p_value} shows that distribution of 2D CNN results is statistically significantly different from the rest with p-values below $0.05$.
Therefore,
2D CNN was used as the path feature extractor as seen in 
\autoref{fig:policy_net_arch}.

A limitation of this analysis approach is the need to assume that the results are independent.
This is of course not necessarily true as changing the feature extractor might change the requirements for $N_\text{width}$. Furthermore,
the high p-value score for the path feature extraction method implies that more runs could have been undertaken.
It may also imply that this value is not important to the network.
However,
a more complex analysis is outwith the scope of this study.

\newcommand{\imgwidth}{0.45\linewidth}
\begin{figure}[htbp]
	\centering
	\subfloat[$N_\text{width}$\label{fig:probability_efficiency_mean_vs_layer_width}]{
		\includegraphics[width=\imgwidth]{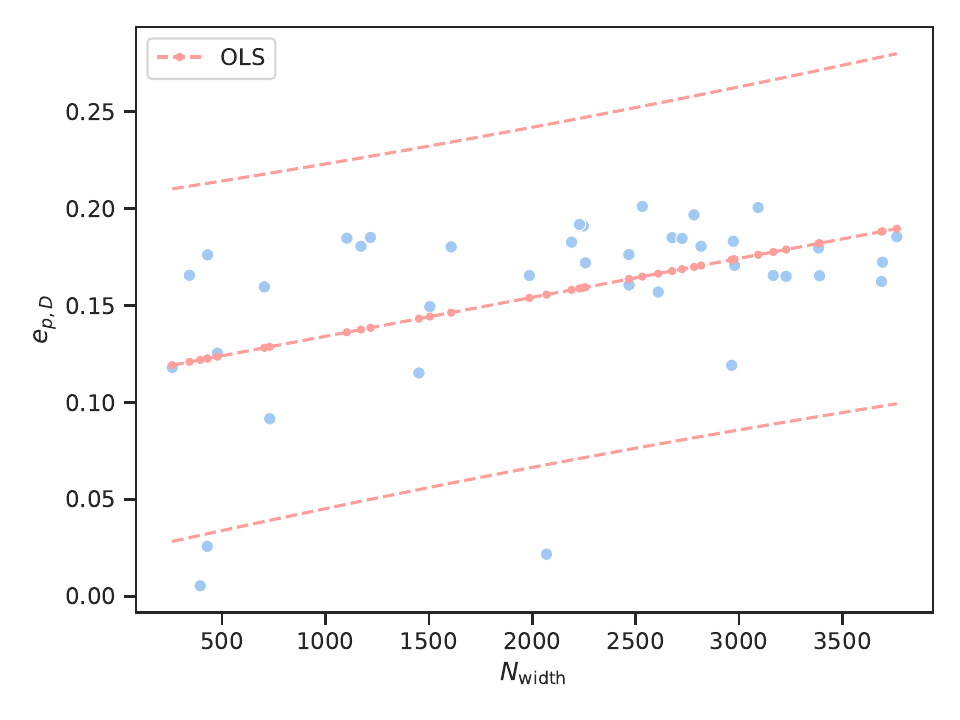}
	}
	\hfill
	\subfloat[$N_\text{layer}$\label{fig:probability_efficiency_mean_vs_n_layers}]{
		\includegraphics[width=\imgwidth]{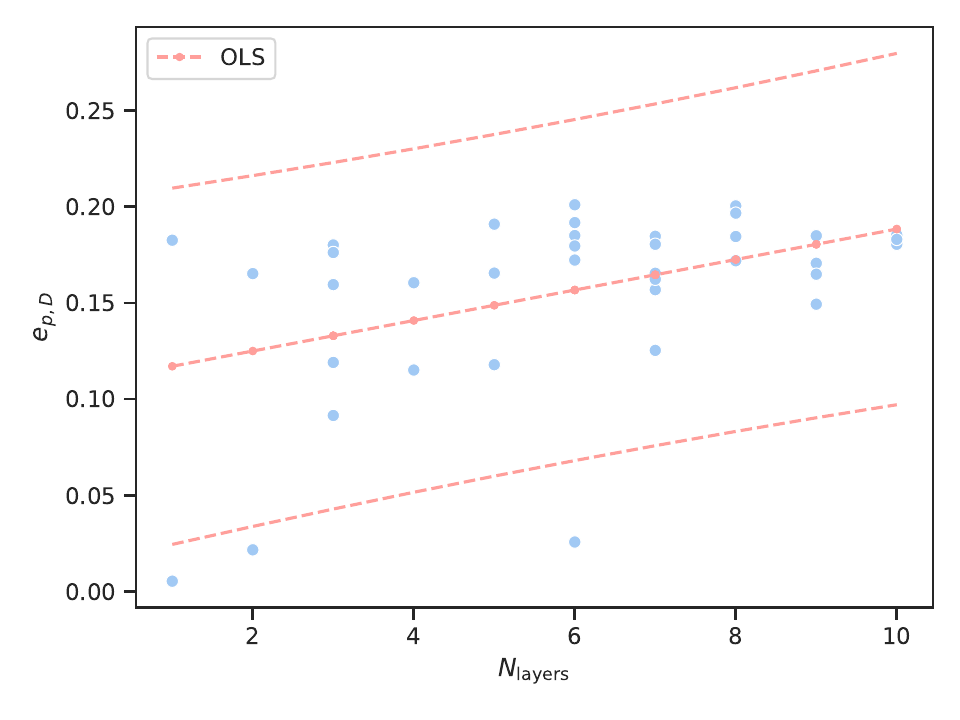}
	}
	\vfill
	\subfloat[Path feature extraction method\label{fig:probability_efficiency_mean_vs_path_feature_extraction}]{
		\includegraphics[width=\imgwidth]{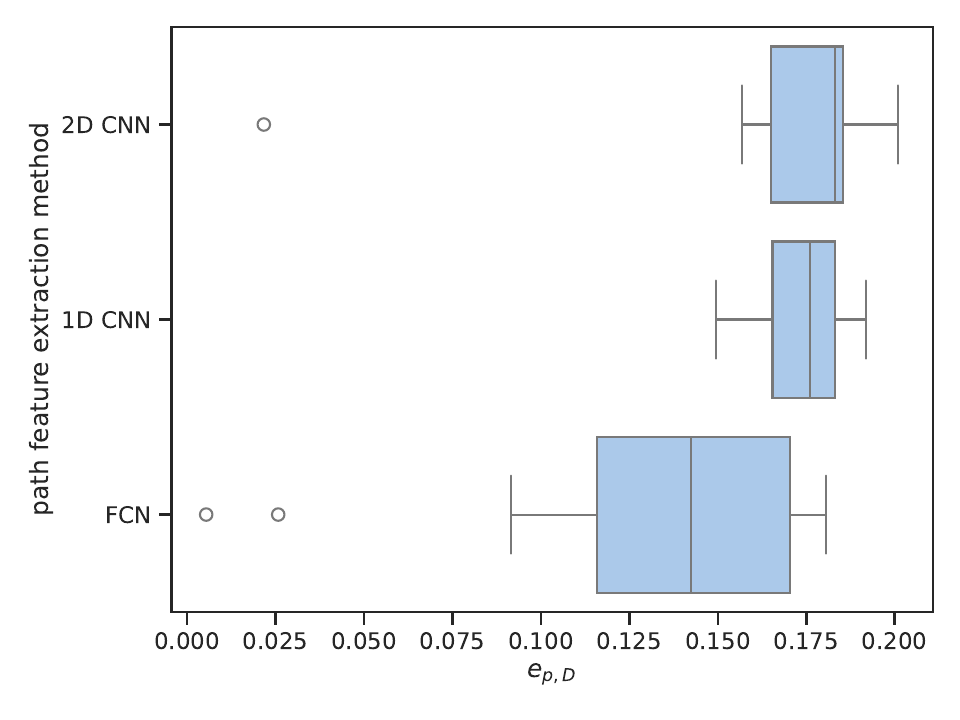}
	}
	\hfill
	\subfloat[Path feature extraction method outptu dimension\label{fig:probability_efficiency_mean_vs_path_output_dim}]{
		\includegraphics[width=\imgwidth]{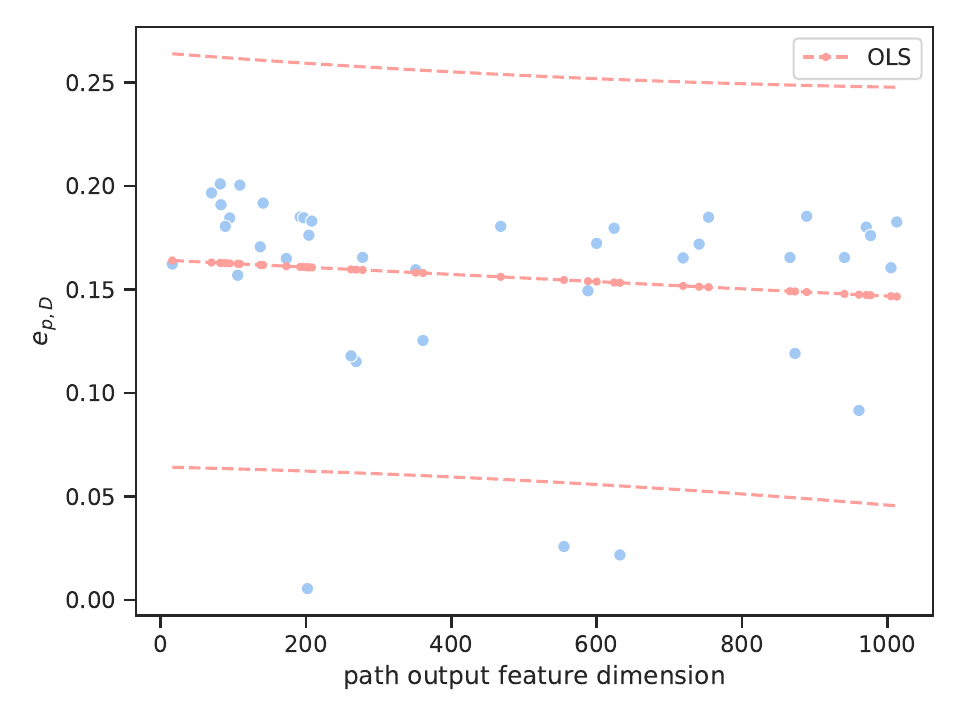}
	}
	\caption{Ordinary least squares (OLS) linear regression fitted to the $e_{p,D}$ results from the hyperparameter sweep.}
	\label{fig:hyperparameter_sweep_ols}
\end{figure}

\begin{figure}[htbp]
	\centering
	\includegraphics[width=0.45\linewidth]{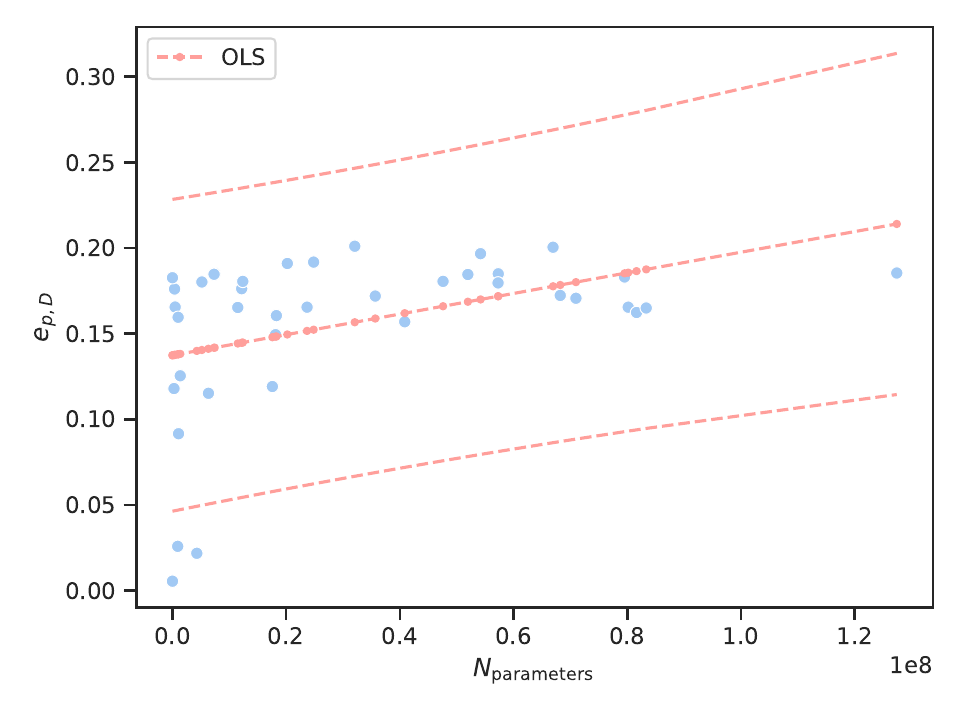}
	\caption{$e_{p,D}$ with the number of parameters within the policy network.
		This value was derived from 
		\autoref{fig:probability_efficiency_mean_vs_layer_width} and 
		\autoref{fig:probability_efficiency_mean_vs_n_layers}.}
	\label{fig:probability_efficiency_mean_vs_n_params}
\end{figure}

